\documentclass[sigconf]{acmart}
\AtBeginDocument{%
  \providecommand\BibTeX{{%
    \normalfont B\kern-0.5em{\scshape i\kern-0.25em b}\kern-0.8em\TeX}}}

\setcopyright{iw3c2w3}
\copyrightyear{2020}
\acmYear{2020}
\acmConference[WWW '20]{Proceedings of The Web Conference 2020}{April 20--24, 2020}{Taipei, Taiwan}
\acmBooktitle{Proceedings of The Web Conference 2020 (WWW '20), April 20--24, 2020, Taipei, Taiwan}
\acmPrice{}
\acmDOI{10.1145/3366423.3380154}
\acmISBN{978-1-4503-7023-3/20/04}

\usepackage{hyperref}       
\usepackage{soul}
\usepackage{booktabs}       
\usepackage{amsfonts}       
\usepackage{nicefrac}       
\usepackage{microtype}      
\usepackage{amsmath,bm}
\usepackage{amsthm}
\usepackage{enumerate}
\usepackage{multirow}
\newcommand{\bx}{{\bm{x}}}

\newcommand{\bD}{{\mathcal{D}}}
\newcommand{\tr}{{\textrm{train}}}
\newcommand{\val}{{\textrm{dev}}}
\newcommand{\te}{{\textrm{test}}}
\newcommand{\phat}{\hat p}

\newcommand{\s}{\sigma}

\begin{document}

\title{Field-aware Calibration: A Simple and Empirically Strong\\ Method for Reliable Probabilistic Predictions}


\author{Feiyang Pan}
\affiliation{
  \institution{Institute of Computing Technology, Chinese Academy of Sciences}
}
\authornote{\,Correspondence to: Feiyang Pan <panfeiyang@ict.ac.cn>.}
\authornote{At Key Lab of Intelligent Information Processing of Chinese Academy of Sciences. Also at University of Chinese Academy of Sciences, China. }

\author{Xiang Ao}
\affiliation{%
\institution{Institute of Computing Technology, Chinese Academy of Sciences}
}
\authornotemark[2]

\author{Pingzhong Tang}
\affiliation{%
 \institution{IIIS, Tsinghua University}
}

\author{Min Lu}
\affiliation{%
 \institution{Tencent}
}

\author{Dapeng Liu}
\affiliation{%
 \institution{Tencent}
}

\author{Lei Xiao}
\affiliation{%
 \institution{Tencent}
}

\author{Qing He}
\affiliation{%
 \institution{Institute of Computing Technology, Chinese Academy of Sciences}
}
\authornotemark[2]

\renewcommand{\shortauthors}{Pan, et al.}
\begin{abstract}
It is often observed that the probabilistic predictions given by a machine learning model can disagree with averaged actual outcomes on specific subsets of data, which is also known as the issue of {\em miscalibration}. It is responsible for the unreliability of practical machine learning systems. For example, in online advertising, an ad can receive a click-through rate prediction of 0.1 over some population of users where its actual click rate is 0.15. In such cases, the probabilistic predictions have to be fixed before the system can be deployed.

In this paper, we first introduce a new evaluation metric named {\em field-level calibration error} that measures the bias in predictions over the sensitive input field that the decision-maker concerns. We show that existing post-hoc calibration methods have limited improvements in the new field-level metric and other non-calibration metrics such as the AUC score. To this end, we propose Neural Calibration, a simple yet powerful post-hoc calibration method that learns to calibrate by making full use of the field-aware information over the validation set. We present extensive experiments on five large-scale datasets. The results showed that Neural Calibration significantly improves against uncalibrated predictions in common metrics such as the negative log-likelihood, Brier score and AUC, as well as the proposed field-level calibration error.
\end{abstract}



\maketitle
\section{Introduction}
To make reliable decisions, it is at the heart of machine learning models to provide accurate probabilistic predictions. Unfortunately, recent studies have observed that many existing machine learning methods, especially deep learning methods, can yield poorly calibrated probabilistic predictions~\cite{korb1999calibration,bella2010calibration,guo2017calibration,ovadia2019can}, which hurts reliability of the decision-making systems.

A predictive machine learning model is said to be well-calibrated if it makes probabilistic predictions or likelihood estimations that agree with the actual outcomes~\cite{naeini2015obtaining,guo2017calibration}. That is, when the model makes predictions on an unseen dataset, in any subset of the data, if the averaged likelihood estimation is $p$, the actual outcomes do occur around $p$ fraction of the times. For example, in online advertising, when a well-calibrated machine learning model predicts a click-through rate of $10\%$ for an ad, the ad does being clicked at 10\% of the times.

However, it is hard to obtain well-calibrated probabilistic predictions. Recent studies~\cite{guo2017calibration,borisov2018calibration,geifman2018bias} reported that, in the fields of computer vision and information retrieval, deep neural networks are  poorly calibrated, i.e., a model can ``confidently'' make mistakes. Miscalibration not only affects the overall utility, but also undermines the reliability over certain groups of data, e.g., a model can overestimate the click rate for one advertisement but underestimate the click rate for another. In particular, machine learning models especially those with deep and complex structures can make desirable predictions in terms of non-calibration performance measures, but suffer from poor results in calibration-related measures~\cite{guo2017calibration}.

Let us look at an example, the details of which are shown in Section~\ref{sec:obs}. We train a multi-layer perceptron (MLP) to predict whether an issued loan would be in default over a public dataset\footnote{Lending Club dataset: https://www.kaggle.com/wendykan/lending-club-loan-data}. The trained neural network works nicely with a high AUC 
score on the test set, which indicates that the model is probably ``good'' to be used to make decisions. Unfortunately, when we look into fine-grained results over borrowers in different states in the U.S., we found the model unreliable: It overestimates the defaulter probabilities for some states but underestimates the others. In this case, the trained model seems ``powerful'' for certain utility functions such as the AUC score, but is unreliable because of the large miscalibration error on some specific subsets of data. Therefore, it is crucial to calibrate the predictions so as not to mislead the decision-maker.

For this purpose, we first raise the following question.

{\it \textbf{Q1}. Given probabilistic predictions on a test dataset, how to measure the calibration error?}%

Perhaps surprisingly, standard metrics are insufficient to evaluate miscalibration errors. Negative Log-Likelihood and the Brier score, arguably the two most popular metrics, can solely evaluate the error on instance level, which is too fine-grained to measure miscalibration on subsets. In particular, they cannot be used to report biases over subsets of data. On the other hand, the reliability diagram~\cite{degroot1983comparison} can only visualize the averaged the error on probability intervals, thus might be too coarse-grained at the subset level.

To answer {\it \textbf{Q1}}, we introduce a new class of evaluation metrics, coined the \emph{Field-level Calibration Error}. It can evaluate the averaged bias of predictions over specific input fields, which is especially useful on categorical data. Take the loan defaulter prediction task as an example. The new metric can measure the field-level error on input such as ``state''.
We observe that the proposed field-level calibration error indeed measures the error ignored by previous metrics. That is, a set of predictions can simultaneously get a high AUC score, low Log-loss, but large field-level calibration error.

To fix miscalibration, various methods have been proposed, e.g., \cite{Isotonic-barlow1972statistical,platt1999probabilistic,zadrozny2002transforming,niculescu2005predicting,naeini2015obtaining,guo2017calibration}. A standard pipeline for calibration requires a development (validation) dataset to learn a mapping function that transforms raw model outputs into calibrated probabilities. By using such post-hoc calibration, the error on the hold-out data can then be reduced, while the order of outputs in the test set can keep unchanged. However, it is reported \cite{ovadia2019can} that these methods are unreliable under dataset shift.

Further, machine learning practitioners might immediately find such methods suboptimal: Recall that there is a development dataset, we can directly train a model over the joint of the training set and the development set, which may probably result in a better model. So why do we bother splitting the data and doing the post-hoc calibration? We observed that training a model over the whole dataset can indeed reach a much higher AUC score than the mentioned post-hoc calibration methods, but is unreliable because sometimes the predictions can be poorly calibrated.

In light of these observations, a practical question arises:

{\it \textbf{Q2}. Is there any way to simultaneously reduce the calibration error and improve other non-calibration metrics such as the AUC score?}%

\begin{figure}[!t]
\centering
\includegraphics[width=\columnwidth]{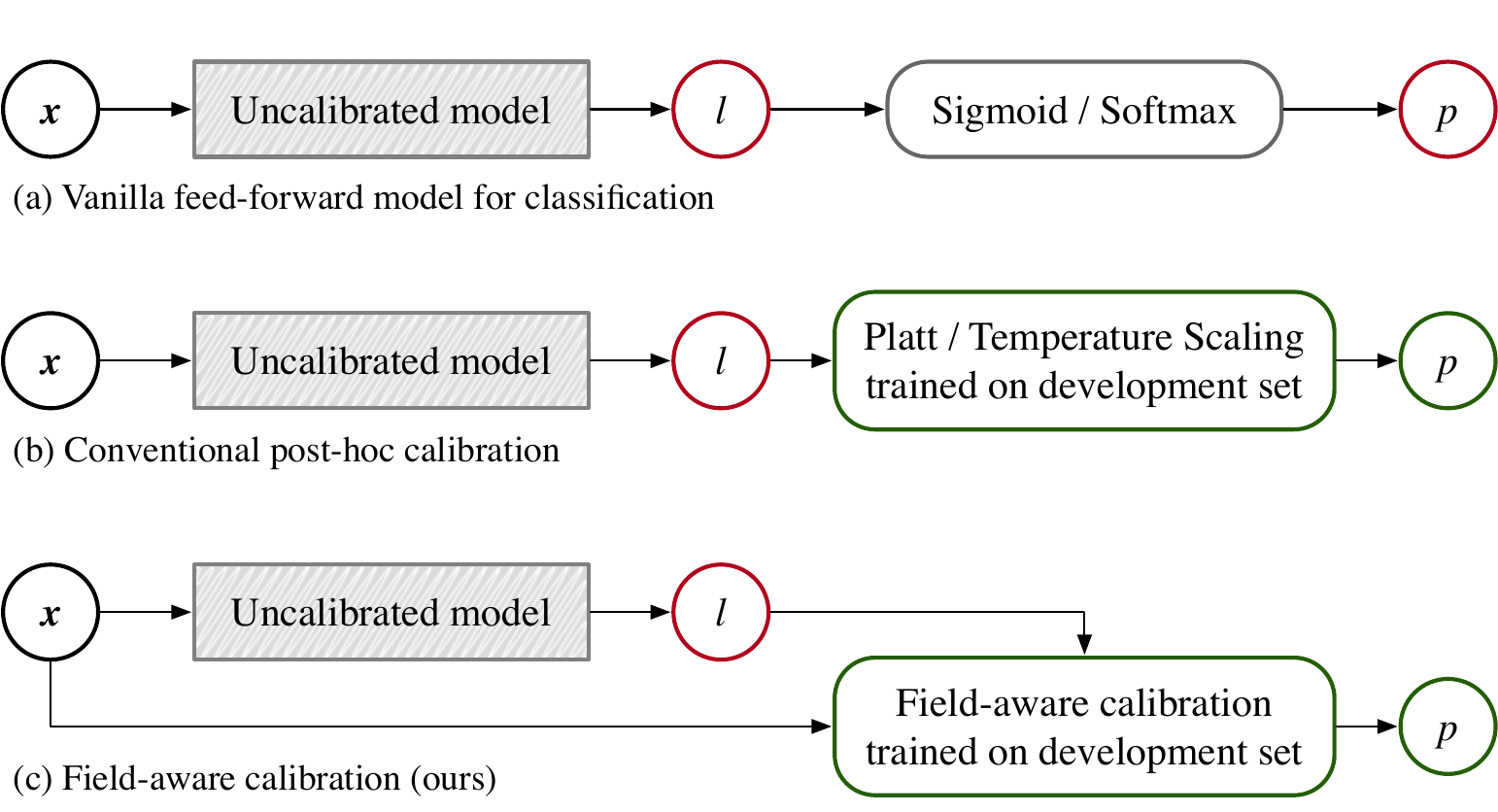}
\caption{Comparison of existing pipelines and ours.}
\label{fig-one}
\end{figure}

Our answer is ``Yes''. To achieve this, we propose a method based on neural networks, coined Neural Calibration. Different from the mentioned methods, Neural Calibration is a field-aware calibration method. The training pipeline of Neural Calibration and comparisons with existing methods are demonstrated in Figure \ref{fig-one}. Rather than learning a mapping from the raw model output to a calibrated probability, 
Neural Calibration trains a neural network over the validation set by taking both the raw model output and all other features as inputs. 

We also study the effect of dataset shift to various calibration methods. We empirically find that common post-hoc calibration methods, including Platt Scaling~\cite{platt1999probabilistic,niculescu2005predicting} and Isotonic Regression~\cite{Isotonic-barlow1972statistical,niculescu2005predicting}, can be unreliable under data shift. When the distribution of development set is not similar to the test set, the calibrated probabilities might be worse than the raw outputs. On the contrary, our proposed Neural Calibration can be empirically robust and reliable, even when the development set is not close to the test set.

We conducted extensive experiments over five large scale real-world datasets to verify the effectiveness. We show that by using our learning pipeline, the final predictions can not only achieve lower calibration error than previous calibration methods but also reach a comparable or better performance on non-calibration metrics compared with the joint training pipeline.

Our contribution can be summarized as follows:
\begin{itemize}
\item We put forward Field-level Calibration Error, a new type of metric for probabilistic predictions. It focuses on measuring miscalibration on specific subsets of data 
that are overlooked by existing metrics.

\item We propose Neural Calibration, a parametric field-aware calibration method, which makes a full use of the development data. The proposed method is easy to implement yet empirically powerful.

\item It follows a pipeline for practitioners in machine learning and data mining, which can improve the uncalibrated outputs in both calibration metrics and non-calibration metrics. We observed through extensive experiments that Neural Calibration could overcome several drawbacks of existing methods and achieve significantly better performance.
\end{itemize}

\section{Preliminary}
This paper focuses on calibrating probabilistic predictions on classification tasks with binary outcomes. The probabilistic prediction to estimate is $p(\bx)=\Pr(Y=1\mid \bx)$ where $\bx$ is the input and $Y\in\{0,1\}$ is the random variable of the binary outcome. This probability can be used to quantify the likelihood of data, the uncertainty on a data point, or the model's confidence on its prediction.

 Modern deep learning models obtains the probabilistic prediction by a base model with a $\mathsf{sigmoid}$ output
\begin{equation}\phat(\bx) = \s(l) = \s(f(\bx))\end{equation}
where $\s(\cdot)$ denotes the $\mathsf{sigmoid}$ function and $l=f(\bx)$ is the uncalibrated non-probabilistic output~(also known as the \textit{logit}) of the discriminative model.

We denote a labeled dataset as $\bD=\{(\bx_{i}, y_{i})\}_{i=1}^{|\bD|}$. The training / development (validation) / test set are denoted by $\bD_{\tr}, \bD_{\val}, \bD_{\te}$, respectively. Note that we do not assume that the three datasets are from the same distribution, that is, we allow for minor data-shift among the datasets. For the simplicity of notations, we will use $\phat_{i}$ to denote $\phat(\bx_{i})$.
\subsection{Existing metrics for probabilistic predictions}\label{sec:metrics}
\subsubsection{Instance-level calibration error}\label{sec:instance-ece}
A straight-forward way to measure miscalibration is to average the errors over all the instances in a dataset. For example, Negative Log-Likelihood (NLL), also known as the Log-loss, is formulated as
\begin{equation}
\textrm{LogLoss} = \frac{1}{|\bD|}\sum_{i=1}^{|\bD|}\big[-y_{i}\log \phat_{i} -(1-y_{i})\log(1-\phat_{i})\big].
\end{equation}
The Brier score is the mean squared error over instances
\begin{equation}
\textrm{BrierScore} = \frac{1}{|\bD|}\sum_{i=1}^{|\bD|}\big(y_{i}-\phat_{i}\big)^{2}.
\end{equation}
Therefore, by minimizing these two metrics, the overall bias becomes smaller.
However, a drawback for these two metrics is that they cannot measure the bias in groups of instances. Their value is not easy to interpret, e.g., by knowing a Log-loss of 0.2, one cannot even guess the averaged bias of the predictions intuitively.

\subsubsection{Probability-level calibration error}\label{sec:prob-ece}
In many previous studies \cite{degroot1983comparison,naeini2015obtaining,guo2017calibration}, the calibration error is formulated by partitioning the predictions into bins and summing up the errors over the bins. Formally, we partition the $[0,1)$ interval into $K$ partitions where the $k^{\textrm{th}}$ interval is $[a_{k}, b_{k})$. Recall that we discuss the case of binary classification, the error is
\begin{equation}
\textrm{Prob-ECE} = \frac{1}{|\bD|}\sum_{k=1}^{K}\big\vert \sum_{i=1}^{|\bD|} (y_{i}-\phat_{i})\mathbf{1}_{[a_{k}, b_{k})}(\phat_{i} )\big\vert,
\label{PECE-true}\end{equation}
where $\mathbf{1}_{[a_{k}, b_{k})}(\phat_{i})$ is an indicator function which equals to 1 if $\phat_{i}\in[a_{k}, b_{k})$. For this metric, the goal can be understood as ``for every subset of data where the predicted positive rate is $p$, the real rate should be around $p$''. A reliability diagram showing the probability-level error is shown on the left side of Figure \ref{fig-metrics}.

However, this metric is somewhat rough for evaluating predictions of specific tasks where the labels are highly imbalanced. Also, it might be misleading for real-world applications such as click-through rate prediction in online advertising. For example, if a model predicts an averaged click-through rate over an advertising platform at every query it sees, the system achieves perfect Prob-ECE but is not so useful in practice. Therefore, this paper does not include Prob-ECE as an evaluation metric.

\subsubsection{Non-calibration metrics}
Many non-calibration metrics can also be used to evaluate probabilistic predictions, such as the classification accuracy, the precision/recall rate, the F-scores, and the area under the receiver operating characteristic curve (AUC).

Among them, the classification accuracy, precision, recall, and the F-scores all require a hard threshold to classify the predictions into hard categorical outputs, while the AUC score does not require a threshold and is more of a measure for the order of predictions. Therefore, the former metrics are often suitable for tasks such as image classification, where the ground-truth label for each instance is explicitly known. On the other hand, the AUC score is preferred in the tasks where the labels are noisy and imbalanced such as user response prediction on the web. It does not require a threshold and is irrelevant to the scale of the predictions, i.e., if we reduce all predictions by 10\%, the AUC score stays the same. Therefore, in this paper, we use the AUC score as the main non-calibration metric.

\subsection{Related work}\label{sec:existing-cali}
Many applications rely on good estimates of predictive uncertainty or data likelihood, including active learning \cite{gal2017deep}, no-regret online learning \cite{riquelme2018deep,pan2019pgcr}, and reinforcement learning \cite{burda2018large,pan2019policy}. Generally speaking, people observe miscalibration when using a model to make inferences on an unseen dataset, which then becomes a development dataset that can be used to calibrate the model. Therefore, it is necessary to fix the error by learning a calibration function using the new development set. Such a two-step method is named post-hoc calibration.

Existing post-hoc calibration methods can be categorized into non-parametric and parametric methods based on the type of mapping functions. Non-parametric methods include binning methods~\cite{zadrozny2002transforming,naeini2015obtaining} and Isotonic Regression~\cite{Isotonic-barlow1972statistical,niculescu2005predicting}. Binning methods partition the raw prediction into bins, each assigned with the averaged outcomes of validation instances in this bin. If the partitions are predefined, the method is known as Histogram Binning~\cite{zadrozny2002transforming}. However, such a mapping cannot keep the order of predictions, thus cannot be used for applications, including advertising. So Isotonic Regression~\cite{Isotonic-barlow1972statistical}, which requires the mapping to be non-decreasing, is widely used in real-world industry~\cite{mcmahan2013ad,borisov2018calibration}.
Parametric methods, on the other hand, use parameterized functions as the mapping function. The most common choice, Platt Scaling~\cite{platt1999probabilistic,niculescu2005predicting}, is equivalent to a univariate logistic regression that transforms the model output~(the logit) into calibrated probabilities. Because of the simple form, Platt scaling can extend to multi-class classification for image and text classification~\cite{guo2017calibration}. However, such oversimplified mapping tends to under-fit the data and might be sub-optimal.

In the industry, calibration is a basic component in large-scale information retrieval systems. For example, in the click-through rate prediction systems of advertising~\cite{graepel2010web,mcmahan2013ad,he2014practical} and search~\cite{borisov2018calibration}, calibration plays a critical role. Platt Scaling~\cite{graepel2010web,mcmahan2013ad} and Isotonic Regression~\cite{he2014practical,borisov2018calibration} are commonly used. Our paper introduces a simple novel method that is more powerful than these baselines.

Other related work includes calibration with more detailed mapping functions or in different problem settings. To name some, \cite{naeini2015obtaining} extends Histogram Binning to a Bayes ensemble; \cite{guo2017calibration} extends Platt scaling to Temperature scaling; \cite{neelon2004bayesian} extends Isotonic Regression to Bayesian, \cite{kotlowski2016online} generalizes it in an online setting, \cite{borisov2018calibration} uses it for calibrating click models; and \cite{lakshminarayanan2017simple} uses model ensemble to reduce the bias of predictions of deep learning models. The most related concept of previous work is group-wise calibration~\cite{crowson2016assessing}, which considers fixing the error across various groups of data. However, their method is solely a linear model over the considered groups and cannot be applied to large-scale problems. The field-level evaluation criteria proposed in this paper is also discussed in~\cite{kleinberg2016inherent,pmlr-v80-hebert-johnson18a}. However, \cite{kleinberg2016inherent} does not give an explicit formulation of the evaluation metric or a method to fix the error, and the method of~\cite{pmlr-v80-hebert-johnson18a} is computationally expensive that needs to iterate every subset of data. Different from the previous work, we give explicit formulation for the metric as well as an efficient algorithm to calibrate it.

There is another line of work studying the reasons why miscalibration occurs, or how to alleviate miscalibration during training the base model~\cite{degroot1983comparison,hendrycks2016baseline,guo2017calibration,lakshminarayanan2017simple,lee2017training}.
This paper studies the metrics to evaluate miscalibration and a practical method to fix it. The term ``calibration'' is also often used in the literature of fairness-aware classification~\cite{kamishima2011fairness,zafar2015fairness,kleinberg2016inherent,pleiss2017fairness,menon2018cost,pmlr-v80-hebert-johnson18a}. In fairness literature, it aims to give fair predictions for specific features, e.g., a fair model should predict the same acceptance rates for female and male applicants, which is different from the target of this paper. 

\section{Field-level calibration error}
\begin{figure}[!tb]
\centering
\includegraphics[width=\columnwidth]{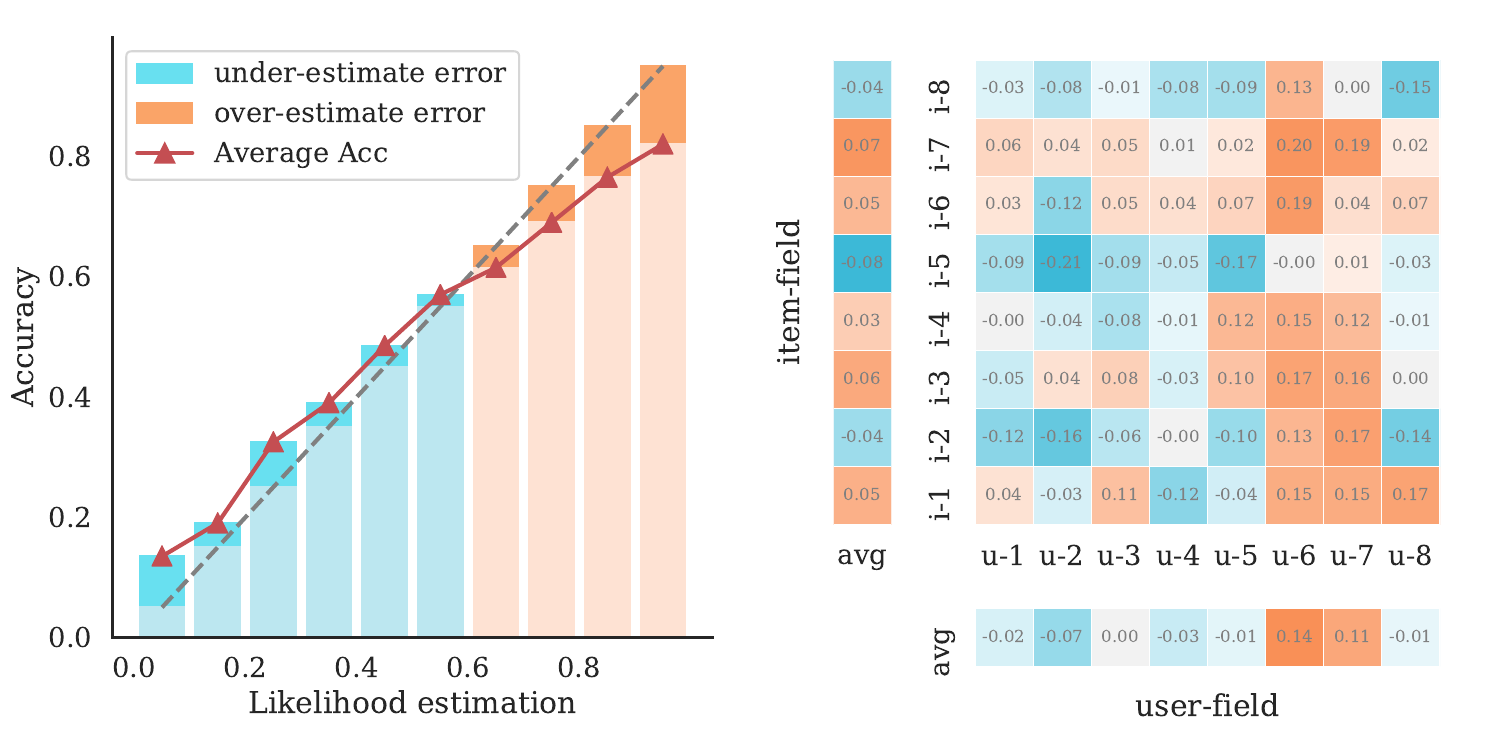}
\caption{A sketch of evaluation metrics of the calibration error on the test set. Left: A reliability diagram, which visualizes the probability-level calibration error. The gaps between the red line (actual outcome) and the dashed line (predictions) are the calibration errors in the intervals. Right: Field-level calibration errors on two fields: user-field and item-field, in a recommender system where one needs to predict the probability of a user likes an item. The numbers in the blocks are the difference between the predicted likelihood and the real response.}
\label{fig-metrics}
\end{figure}
\begin{figure*}[!t]
\centering
\includegraphics[width=0.9\textwidth]{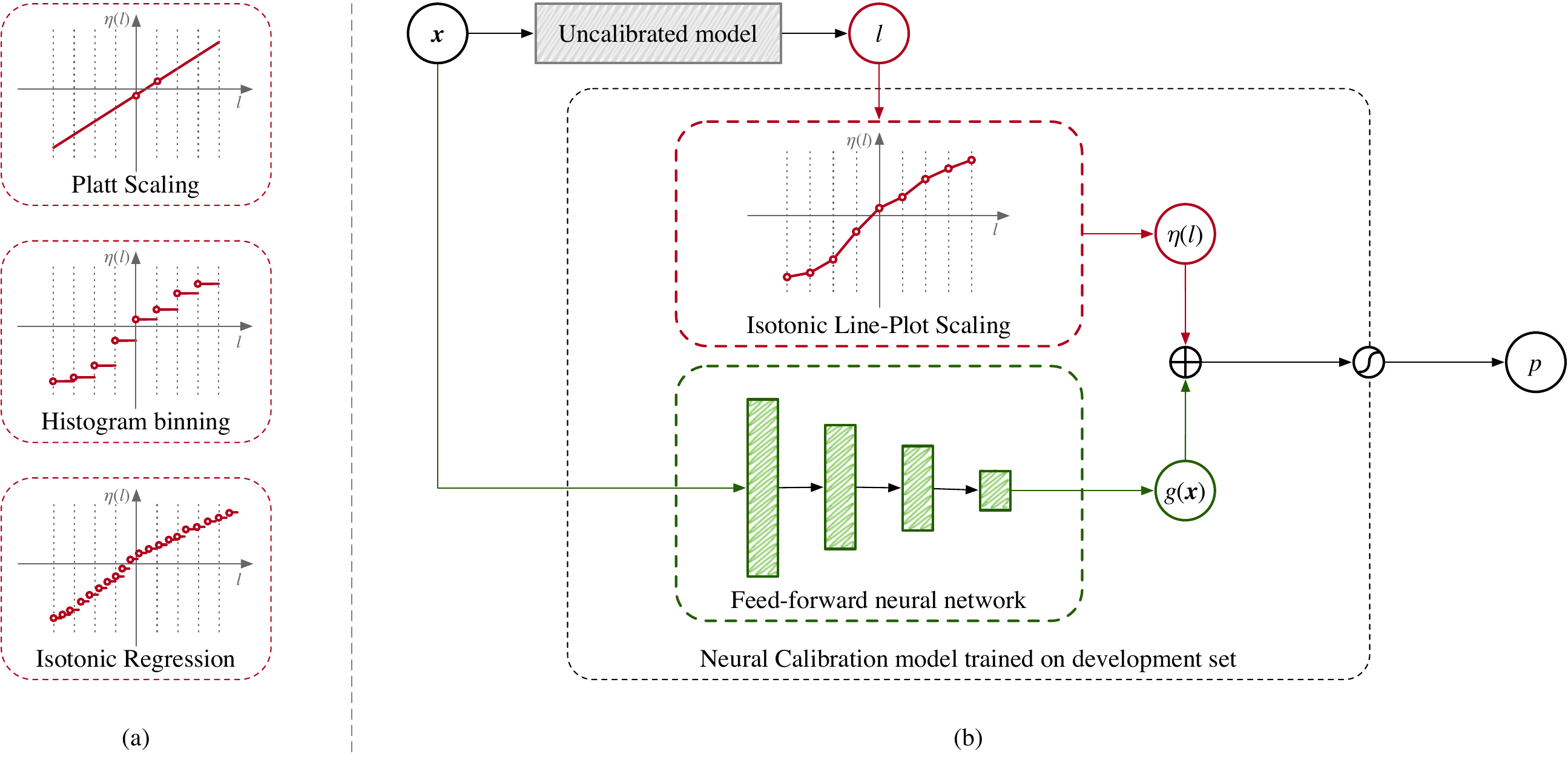}
\caption{(a) Traditional calibration methods that take only the uncalibrated logit $l$ as input. (b) Neural Calibration that takes the original inputs $\bx$ and the uncalibrated logit $l$ as inputs and outputs the calibrated probability. The calibration model on the right-hand side can be trained in an end-to-end manner because the functions are differentiable almost everywhere.}
\label{fig-model}
\end{figure*}

We put forward the field-level calibration error as a new metric to evaluate the reliability of probabilistic predictions. In particular, it measures the biases of predictions in different subsets of the dataset, which is especially vital in applications on the web. A sketch for our field-level metrics is shown on the right side of Figure \ref{fig-metrics}.

Consider the input of the classification problem is a $d+1$ dimensional vector $\bx =(z, x_{1},\dots, x_{d})$ including one specific categorical field $z\in\mathcal{Z}$ that the decision-maker especially cares about. Given that $z$ is a categorical feature, we can partition the input space into $|\mathcal{Z}|$ disjoint subsets. For example, in the loan defaulter prediction task mentioned previously, this particular field is the ``address state'' feature with 51 levels, i.e., $z\in\mathcal{Z}=\{1,\dots,51\}$. Thus the data can be partitioned into 51 disjoint subsets.

Now we use these subsets to formulate field-level calibration errors. We denote the event of an instance belongs to a subset by $\mathbf{1}_{[z_{i}=z]}$ whose value is either 0 or 1. We then formulate the field-level expected calibration error (Field-ECE or F-ECE) as a weighted sum of the averaged bias of predictions in each subset
\begin{equation}
\textrm{Field-ECE} = \frac{1}{|\bD|}\sum_{z=1}^{|\mathcal{Z}|}\big\vert \sum_{i=1}^{|D|} (y_{i}-\phat_{i})\mathbf{1}_{[z_{i}=z]} \big\vert,
\label{FECE}\end{equation}
which is straight-forward: ``for every subset of data categorized by the field $z$, the averaged prediction should agree with the averaged outcome''. Therefore, if a set of predictions gets a large Field-ECE, it indicates that the predictions are biased on some part of the data.

Although this formulation has a similar form to Prob-ECE, there is a crucial difference. In Prob-ECE, the partition is determined by the prediction $\phat$ itself so that the result can be misleading, e.g., it can get nearly perfect Prob-ECE by predicting a constant because all the instances go into one bin. But in our Field-ECE, the partition is determined by the input feature $z$, so the resultant metric can be consistent without being affected.

Further, we can have the field-level relative calibration error formulated as the averaged rate of errors divided by the true outcomes,
 \begin{equation}
\textrm{Field-RCE} = \frac{1}{|\bD|}\sum_{z=1}^{|\mathcal{Z}|}N_z\frac{\big\vert \sum_{i=1}^{|\bD|} (y_{i}-\phat_{i})\mathbf{1}_{[z_{i}=z]} \big\vert}{\sum_{i=1}^{|\bD|} (y_{i}+\epsilon)\mathbf{1}_{[z_{i}=z]}},
\label{FRCE}\end{equation}
where $N_z$ is the number of instance in each subset, i.e., $\sum_{z=1}^{|\mathcal{Z}|} N_z=|\bD|$, and $\epsilon$ is a small positive number to prevent division by zero, e.g., $\epsilon=0.01$. We suggest not to set $\epsilon$ too small in order to reduce the variance of the result.

Note that these measures are proposed as evaluation metrics on test data. It is not proper to use them as loss functions on the training set. Also, we suggest to use them along with other common metrics such as the Log-loss and the Brier score, because the field-level and instance-level metrics measure errors at different scales. We also want to emphasize that although field-level calibration errors are formulated upon a categorical input field $z$, they can be easily extended to non-categorical fields by discretizing them into disjoint subsets. 

\section{Neural Calibration}

We are motivated to design a new method that can improve both calibration and non-calibration metrics. Our proposed solution is named Neural Calibration. It consists of two modules: 1) a parametric univariate mapping function to transform the original model output into a calibrated one, and 2) an auxiliary neural network to fully exploit the development set. The overall architecture of Neural Calibration is shown on the right-hand side of Figure~\ref{fig-model}.

The basic formulation is written as follows
\begin{equation}
q(l, \bx) = \s\big( \eta(l) + g(\bx) \big)
\end{equation}
where $\s(\cdot)$ denotes for a sigmoid function. The input of the sigmoid function is the sum of two sub-modules: $\eta(l)$ transforms the uncalibrated logit $l=f(\bx)$ given by the original model into a calibrated one, and $g(\bx)$ is an auxiliary neural network for field-aware calibration that takes the raw features $\bx$ as inputs.

Therefore, there are two functions $\eta(\cdot)$ and $g(\cdot)$ to learn, which are parametrized with trainable parameters $\phi$ and $\theta$, respectively. They need to be trained simultaneously over the validation set.

The objective for training Neural Calibration is to minimize the Log-loss over the validation set, i.e.,
\begin{equation}
\min_{\phi,\theta} \frac{1}{|\bD_{\val}|}\sum_{i=1}^{|\bD_{\val}|}\big[-y_{i}\log q_{i} - (1-y_{i})\log(1-q_{i})\big].\label{nc-obj}
\end{equation}
Therefore, Neural Calibration can be trained by stochastic gradient descent, just like any other deep learning models.

Such a parametric method has many benefits. It can be updated by mini-batch gradient descent, so it is easy to deal with streaming data in an online setting. Also, it is flexible that can extend to multi-class classification without much difficulty.

Now we introduce the detailed function structure of the two modules $\eta(l)$ and $g(\bx)$.

\subsection{Isotonic Line-Plot Scaling (ILPS)}
We are interested in finding a stronger $\eta(\cdot)$ that achieves high fitting power as well as some nice properties.
To enhance stronger fitting power, we borrow the spirit of binning from non-parametric methods. To enable efficient online training, we make it a parametric function so that the whole model can be updated by gradient descent. To make it reliable and interpretable, we use a sophisticated structure and constraints to make it an isotonic~(non-decreasing) continuous piece-wise linear function. The difference between ILPS from conventional methods is demonstrated in Figure~\ref{fig-model}.

We first partition a support set $(M_{1}, M_{2})\subset \mathbb{R}$ into several intervals with fixed splits $M_{1}=a_{1}<a_{2}<\dots<a_{K+1}=M_{2}$ where $M_{1}, M_{2}$ are two pre-specified numbers. Then we design the scaling function as
\begin{equation}
\eta(l) = b + \sum_{k=1}^{K} w_{k} \min\big\{(l-a_{k})_{+}, a_{k+1}-a_{k}\big\},
\end{equation}
where $(l-a_{k})_{+}$ is short for $\max\{l-a_{k}, 0\}$, and $b, w_1, \dots, w_{K}$ are the trainable parameters to learn. This formulation is easy to understand since it is also known as a continuous piece-wise linear function, where $b$ is the bias term and $w_{1},\dots,w_{K}$ are the slope rate of each piece of linear functions. In this way, the scaling function is continuous, and the total number of parameters is $K+1$ which makes the function both expressive and robust.

However, this parameterized function is not easy to tune by gradient descent because every instance that lies in the $k^{\textrm{th}}$ interval involves $k+1$ parameters during back-propagation. To make the optimization problem easier to solve, we further re-parameterize it
\begin{equation}
\eta(l) = \sum_{k=1}^{K} \bigg[b_{k} + (l-a_{k})\frac{b_{k+1}-b_{k}}{a_{k+1}-a_{k}}\bigg] \mathbf{1}_{[a_{k},a_{k+1})}(l),
\end{equation}
where $\mathbf{1}_{[a_{k},a_{k+1})}(l)$ is an indicator function whose value is 1 if $l\in[a_{k},a_{k+1})$ and is 0 otherwise, $b_{k}$ is the bias term at the left point of the $k^{\textrm{th}}$ interval, and $\frac{b_{k+1}-b_{k}}{a_{k+1}-a_{k}}$ is the slope rate of the same interval. In this way, for each instance that lies in the $k^{\textrm{th}}$ interval, it only involves two parameters $b_{k}$ and $b_{k+1}$ during back-propagation which makes the optimization much easier. This mapping function looks just like a line-plot, as it connects the points $(a_{1}, b_{1}),\dots,(a_{K+1}, b_{K+1})$ one by one. The total number in parameters $\bm{b}=(b_{1},\dots,b_{K+1})$ is still $K+1$.

Further, we would like to restrict the function to be isotonic~(non-decreasing) as a calibration mapping should keep the order of non-calibrated outputs. To achieve this, we put a constraint on the parameters, i.e.,
$b_{k}\leq b_{k+1}$, $\forall 1\leq k \leq K$. In the actual implementation, the constraint is realized by the Lagrange method, i.e., adding a regularization term on the loss function so the overall optimization problem can be solved by gradient descent.

Now we get a novel parametric calibration mapping $\eta(\cdot)$ named Isotonic Line-Plot Scaling (ILPS), because it is a non-decreasing, continuous piece-wise linear function - that looks like a line-plot. A sketch of such a mapping function is shown in the red dashed box in Figure \ref{fig-model}. Overall, the training objective for ILPS is as follows
\begin{equation}
\mathcal{L}_{\textrm{ILPS}}(\bm{b};y,l) = \textrm{LogLoss}\big(\eta(l), y\big) + \lambda\sum_{k=1}^{K} (b_{k}-b_{k+1})_{+}
\end{equation}
where the first part evaluates the fitting of the mapping function and the second part is the regularization term to ensure the mapping is non-decreasing. The regularization gets activated only if there exists some interval where $b_{k}>b_{k+1}$.

 In practice, we set $K=100$ and the splits s.t. $\s(a_k)=k/(K+1)$ for $k=1,\dots,K$. We will show in the ablation study~(Section \ref{sec:expresults}) that this mapping can significantly outperform the traditional parametric method Platt scaling and also be comparable or better than common non-parametric methods.

\subsection{The auxiliary neural network \\for field-aware calibration}
The previous part designed a univariate mapping from the original model output to a calibrated one. We further learn to use the whole development data to train an auxiliary neural network.

The neural network $g(\bx)$ learns to fix the field-level miscalibration or biases by using all necessary features from the development set. Our intuition is simple: if we observe a field-level calibration error over the validation set on a specific field $z$, then we should find a way to fix it by learning a function of both the model output and the field $z$.
In practice, a decision-maker probably cares about the biases in more than one field, for example, an advertising system should have a low bias on the user-level, the ad-level, and the advertiser-level. Considering that in many cases, $z$ is a part of the input $\bx$, we can directly learn a function of $\bx$.

We do not restrict the form of neural network for $g(\bx)$. For example, one can use a multi-layered perceptron for general data mining task, a convolution neural network for image inputs, or a recurrent neural network for sequential data.

\subsection{Learning pipeline}
Now we would like to describe how to use Neural Calibration to improve the performance in real-world machine-learning-powered applications.
Suppose that we want to train a model with a set of labeled binary classification data, and the target is to make reliable probabilistic predictions during inference on the unseen data. We put forward the following pipeline of Neural Calibration:

\textbf{Step 1.} Split the dataset at hand into a training set $\bD_{\tr}$ and a development (validation) set $\bD_{\val}$.

\textbf{Step 2.} Train a base model $f(\bx)$ over the training set $\bD_{\tr}$. If necessary, select the model and tune the hyper-parameters by evaluating on the validation set.

\textbf{Step 3.} Make (uncalibrated) predictions on the development set to get a supervised dataset for calibration $\{\bx_{i}, l_{i}, y_{i}\}$. Train a Neural Calibration model $q(l, \bx)=\s(\eta(l) + g(\bx))$ over this data set.

\textbf{Step 4.} Do inference on the hold-out data with two-step predictions: when a query $\bx$ comes, first predict the uncalibrated logit $l=f(\bx)$, then calibrate it $\hat p = q(l, \bx) = \s(\eta(l) + g(\bx))$.

Here we give a brief explanation.

To begin with, \textbf{Step 1} and \textbf{Step 2} are the common processes in supervised learning.
With the uncalibrated deep model $f(\bx)$, one might observe instance-level, probability-level, or field-level miscalibration by examining the predictions on the validation set. Next, we train the calibration model in \textbf{Step 3}, which can be viewed as training a classification neural network with inputs $(x_{i}, l_{i})$ to fit the label $y_{i}$ by minimizing the Log-loss, as shown in Eq. (\ref{nc-obj}). Finally, during inference on an unseen query with input $\bx$, we can get the final prediction by two steps: first, compute the logit by the original model $l=f(\bx)$, and then compute the calibrated prediction by Neural Calibration $q=\sigma(\eta(l)+g(\bx))$.

\subsubsection*{Comparison with existing offline supervised learning pipelines}

Generally, machine learning or data mining pipelines do not consider the miscalibration problem. That is, to directly make inference after \textbf{Step 1} and \textbf{Step 2}, which results in Model-1 as mentioned in the previous section. In such a case, the validation set is merely used for model selection.

Often, one prefers to making full use of the labeled data at hand, including the validation set. So after training $f(\bx)$ on the training set, one can further update the model according to the validation set, which results in the Model-2 as mentioned. Such a training pipeline is useful, especially when the data is chronologically arranged because the validation set contains samples that are closer to the current time. However, this pipeline does not consider the calibration error.

The pipeline of calibration has the same procedure as ours. However, conventional calibration methods solely learn a mapping from uncalibrated outputs to calibrated predictions at \textbf{Step 3}. Our Neural Calibration is more flexible and powerful because it can fully exploit the validation data.

\section{Experiments}
\subsection{Experimental setup}\label{sec:expsetup}
Through experiments, we seek to answer the following questions:

1. What is the drawback of existing post-hoc calibration methods? What is the drawback of training an uncalibrated model over the joint of training and development set?

2. Can Neural Calibration achieve better performance?

3. How much does $\eta(l)$ and $g(\bx)$ helps, respectively? Why using all input features $\bx$ instead of $z$?

4. Is Neural Calibration more robust to dataset shift comparing to existing methods?

To answer these questions, we set up experiments with five binary classification datasets for data mining.
All the datasets were split into three parts: the training set, the development~(validation) set, and the test set. Note that the supervised examples were not i.i.d. in the three datasets. In fact, we allow for data shift among them. The reason is that in real-world applications on the web, there is always a distribution shift. For example, in any e-commerce or advertising platform, the popularity of items, the distribution of users, the averaged click-through rate, and the conversion rate all change over time. Also, we do not assume that the test set is more similar to the development set against the training set or vice versa, because we simply do not know it in practice. So in our experiments, we split the datasets by date if there is an explicit ``date'' column or by sample index otherwise.

The base model of $f(\bx)$ and $g(\bx)$ are neural networks with the same structure which follows \cite{pan2019warm}. Each column of the raw inputs is first embedded into a 256-dimensional dense embedding vector, which is then concatenated together and fed into a multi-layer perceptron with two 200-dimensional ReLU layers.

For all the tested tasks, we split the datasets into three parts: 60\% for training, 20\% for validation, and the other 20\% for testing.

The experiment pipeline goes like follows. Firstly we train the base model using Adam optimizer~\cite{kingma2014adam} to minimize the Log-loss with a fixed learning rate of 0.001 for one epoch. This trained model is called Model-1. Next, on the one hand, we train common calibration models as well as our proposed Neural Calibration on the development set based on the logits predicted by Model-1. For readability, we denote the calibration pipelines as $\bD_\tr\rightarrow\bD_\val$. On the other hand, we incrementally update Model-1 over the samples from the development set for an epoch, which is called Model-2. To ensure the same learning rate with the calibration methods, we restart the Adam optimizer for Model-2 before the incremental training. We denote the training set for Model-2 as $\bD_\tr\cup\bD_\val$.

For testing, we collect the uncalibrated probabilistic predictions directly from Model-1 and Model-2, and the calibrated predictions in two steps by first make inference using Model-1 and then calibrate by the post-hoc methods. We evaluate the predictions with common metrics, including the Log-loss~(negative log-likelihood), the Brier score, the AUC score, as well as our proposed field-level calibration errors Field-ECE and Field-RCE.

\subsection{Datasets} We tested the methods on five large-scale real-world binary classification datasets. For each data, we arbitrarily choose one categorical field as the field $z$.
\begin{itemize}
\item[1.] \textbf{Lending Club} loan data\footnotemark[1], to predict whether an issued loan will be in default, with 2.26 million samples. The data is split by index as we suppose that the data is arranged by time. The categorical field $z$ is set as the ``\texttt{address state}'' with 51 possible values.

\item[2.] \textbf{Criteo} display advertising data\footnote{https://www.kaggle.com/c/criteo-display-ad-challenge}, a click-through rate (CTR) prediction dataset to predict the probability that the user clicks on a given ad. It consists of 45.8 million samples over 10 days and is split by index. The field $z$ is set as an anonymous feature ``\texttt{C11}'' with 5683 possible values.

\item[3.] \textbf{Avazu} click-through rate prediction data\footnote{https://www.kaggle.com/c/avazu-ctr-prediction}. We used the data of the first 10 days with 40.4 million samples. The data is split by date, i.e., 6 days for training, 2 days for development, and 2 days for testing. The field $z$ is set as the ``\texttt{site ID}'' with 4737 possible values.

\item[4.] \textbf{Porto Seguro}'s safe driver prediction data\footnote{https://www.kaggle.com/c/porto-seguro-safe-driver-prediction}, to predict if a driver will file an insurance claim next year. It has 0.6 million samples and is split by index. The field $z$ is an anonymous categorical field ``\texttt{ps\_ind\_03}'' with 12 possible values.

\item[5.] \textbf{Tencent} click-through rate prediction data, which is subsampled directly from the Tencent's online advertising stream. It consists of 100 million samples of click data across 10 days. To simulate a real application setting, we split the data by date, i.e., 6 days for training, 2 days for development~(validation), and 2 days for testing. The field $z$ is set as the advertisement ID with 0.1 million possible values.

\end{itemize}

\begin{table*}[!th]
\centering
\caption{Results on loan defaulter prediction}
\begin{tabular}{lc|cccc|c}
\toprule
Method & Training data & Log-loss $\downarrow$ & Brier score $\downarrow$ & Field-ECE $\downarrow$ & Field-RCE $\downarrow$ & AUC $\uparrow$\\
\midrule
Base (Model-1) & $\bD_{\tr}$ & 1.255 & 0.082 & 0.077 & 49.2\% & 0.909 \\
\midrule
Base (Model-2) & $\bD_{\tr} \cup \bD_{\val}$ &
1.124 & 0.172 & 0.177 & 115.3\% & 0.985 \\
\midrule
Isotonic Reg. & $\bD_{\tr}\rightarrow \bD_{\val}$ &
0.268 & 0.082 & 0.021 & 13.8\% & 0.912\\
Platt Scaling & $\bD_{\tr}\rightarrow \bD_{\val}$ &
0.287 & 0.080 & 0.037 & 23.9\% & 0.909\\
\midrule
Neural Calibration & $\bD_{\tr}\rightarrow \bD_{\val}$ &
{\bf 0.066} & {\bf 0.016} & {\bf 0.018} & {\bf 12.0\%} & {\bf 0.992} \\
\bottomrule
\end{tabular}
\label{table:loan}\end{table*}

\begin{table*}[!th]
\centering
\caption{Results on Criteo click-through rate prediction}
\begin{tabular}{lc|cccc|c}
\toprule
Method & Training data & Log-loss $\downarrow$ & Brier score $\downarrow$ & Field-ECE $\downarrow$ & Field-RCE $\downarrow$ & AUC $\uparrow$\\
\midrule
Base (Model-1) & $\bD_{\tr}$ & 0.4547 & 0.1474 & 0.0160 & 7.46\% & 0.7967\\
\midrule
Base (Model-2) & $\bD_{\tr} \cup \bD_{\val}$ &
0.4516 & {\bf 0.1464} & 0.0167 & 7.08\% & {\bf 0.8001} \\
\midrule
Isotonic Reg. & $\bD_{\tr}\rightarrow \bD_{\val}$ &
0.4539 & 0.1472 & 0.0134 & 6.09\% & 0.7967\\
Platt Scaling & $\bD_{\tr}\rightarrow \bD_{\val}$ &
0.4539 & 0.1472 & 0.0135 & 6.12\% & 0.7967\\
\midrule
Neural Calibration & $\bD_{\tr}\rightarrow \bD_{\val}$ &
{\bf 0.4513} & {\bf 0.1463} & {\bf 0.0094} & {\bf 4.59\%} & {\bf 0.7996} \\
\bottomrule
\end{tabular}
\label{table:criteo}\end{table*}
\begin{table*}[!th]
\centering
\caption{Results on Avazu click-through rate prediction}
\begin{tabular}{lc|cccc|c}
\toprule
Method & Training data & Log-loss $\downarrow$ & Brier score $\downarrow$ & Field-ECE $\downarrow$ & Field-RCE $\downarrow$ & AUC $\uparrow$\\
\midrule
Base (Model-1) & $\bD_{\tr}$ &
0.3920 & 0.1215 & 0.0139 & 12.88\% & 0.7442\\
\midrule
Base (Model-2) & $\bD_{\tr} \cup \bD_{\val}$ &
0.3875 & {\bf 0.1204} & {\bf 0.0120} & 11.17\% & 0.7496 \\
\midrule
Isotonic Reg. & $\bD_{\tr}\rightarrow \bD_{\val}$ &
0.3917 & 0.1216 & 0.0199 & 18.56\% & 0.7442\\
Platt Scaling & $\bD_{\tr}\rightarrow \bD_{\val}$ &
0.3921 & 0.1215 & 0.0165 & 15.18\% & 0.7442\\
\midrule
Neural Calibration & $\bD_{\tr}\rightarrow \bD_{\val}$ &
{\bf 0.3866} & {\bf 0.1202} & {\bf 0.0121} & {\bf 10.91\%} & {\bf 0.7520} \\
\bottomrule
\end{tabular}
\label{table:avazu}\end{table*}
\begin{table*}[!th]
\centering
\caption{Results on Porto Seguro's safe driver prediction}
\begin{tabular}{lc|cccc|c}
\toprule
Method & Training data & Log-loss $\downarrow$ & Brier score $\downarrow$ & Field-ECE $\downarrow$ & Field-RCE $\downarrow$ & AUC $\uparrow$\\
\midrule
Base (Model-1) & $\bD_{\tr}$ &
0.1552 & 0.0351 & 0.0133 & 28.55\% & 0.6244\\
\midrule
Base (Model-2) & $\bD_{\tr} \cup \bD_{\val}$ &
0.1538 & {\bf 0.0349} & 0.0064 & 13.90\% & 0.6245 \\
\midrule
Isotonic Reg. & $\bD_{\tr}\rightarrow \bD_{\val}$ &
0.1544 & {\bf 0.0349} & 0.0021 & 4.47\% & 0.6244\\
Platt Scaling & $\bD_{\tr}\rightarrow \bD_{\val}$ &
0.1532 & {\bf 0.0349} & 0.0020 & 4.30\% & 0.6244\\
\midrule
Neural Calibration & $\bD_{\tr}\rightarrow \bD_{\val}$ &
{\bf 0.1531} & {\bf 0.0349} & {\bf 0.0018} & {\bf 3.66\%} & {\bf 0.6269} \\
\bottomrule
\end{tabular}
\label{table:porto}\end{table*}
\begin{table*}[!th]
\centering
\caption{Results on Tencent click through-rate prediction}
\begin{tabular}{lc|cccc|c}
\toprule
Method & Training data & Log-loss $\downarrow$ & Brier score $\downarrow$ & Field-ECE $\downarrow$ & Field-RCE $\downarrow$ & AUC $\uparrow$\\
\midrule
Base (Model-1) & $\bD_{\tr}$ &
0.1960 & 0.0522 & 0.0145 & 27.12\% & 0.7885\\
\midrule
Base (Model-2) & $\bD_{\tr} \cup \bD_{\val}$ &
{\bf 0.1953} & {\bf 0.0521} & 0.0128 & 24.58\% & {\bf 0.7907} \\
\midrule
Isotonic Reg. & $\bD_{\tr}\rightarrow \bD_{\val}$ &
0.1958 & 0.0522 & 0.0141 & 25.45\% & 0.7884\\
Platt Scaling & $\bD_{\tr}\rightarrow \bD_{\val}$ &
0.1958 & 0.0522 & 0.0142 & 25.72\% & 0.7885\\
\midrule
Neural Calibration & $\bD_{\tr}\rightarrow \bD_{\val}$ &
{\bf 0.1952} & {\bf 0.0521} & {\bf 0.0124} & {\bf 22.87\%} & {\bf 0.7907} \\
\bottomrule
\end{tabular}
\label{table:tencent}\end{table*}
\begin{table*}[!th]
\centering
\caption{Ablation studies. Field-RCE and AUC on the test set are reported.}
\begin{tabular}{l|cc|cc|cc|cc|cc}
\toprule
 & \multicolumn{2}{c|}{\sc Lending Club} & \multicolumn{2}{c|}{\sc Criteo} & \multicolumn{2}{c|}{\sc Avazu} & \multicolumn{2}{c|}{\sc Porto Seguro} & \multicolumn{2}{c}{\sc Tencent} \\
\cline{2-11}
{\sc Method} & F-RCE & AUC & F-RCE & AUC & F-RCE & AUC & F-RCE & AUC & F-RCE & AUC \\
\midrule
Base ($\bD_{\tr}$) & 49.2\% & 0.909 & 7.46\% & 0.7967 & \bf 12.88\% & 0.7442 & 28.55\% & 0.6244 & 27.12\% & 0.7885 \\
\midrule
Hist. Bin (non-param)& \bf 4.1\% & 0.752 & 6.08\% & 0.7966 & 14.39\% & 0.7440 & 3.98\% & 0.6198 & 25.72\% & 0.7874 \\
Isotonic Reg. (non-param)& 13.8\% & 0.909 & 6.09\% & 0.7967 & 18.56\% & 0.7442 & 4.47\% & 0.6244 & \bf 25.45\% & 0.7884 \\
Platt Scaling (parametric)& 23.9\% & 0.909 & 6.12\% & 0.7967 & 15.18\% & 0.7442 & 4.30\% & 0.6244 & 25.72\% & 0.7885 \\

ILPS (parametric) & 12.1\% & 0.909 & \bf 6.05\% & 0.7967 & \bf 14.25\% & 0.7442 & \bf 3.82\% & 0.6244 & 25.72\% & 0.7885 \\
\midrule
Neural Cali. (ILPS \& field $z$ only) & 11.9\% & 0.909 & \bf 4.54\% & 0.7973 & \bf 8.82\% & 0.7462 & 5.29\% & 0.6238 & 23.11\% & 0.7897 \\
\midrule
Base ($\bD_{\val}$ only) & 127.5\% & 0.915 & 6.11\% & 0.7923 & 15.61\% & 0.7440 & 16.86\% & 0.6177 & 24.48\% & 0.7874 \\
ILPS ($\bD_{\val}\rightarrow\bD_{\tr}$) & \bf 8.3\% & 0.786 & 6.04\% & 0.7923 & 16.44\% & 0.7440 & 4.88\% & 0.6177 & \bf 22.76\% & 0.7874 \\
Neural Cali. ($\bD_{\val}\rightarrow\bD_{\tr}$) & \bf 3.2\% & 0.986 & \bf 4.65\% & 0.7976 &  14.51\% & 0.7482 & 4.42\% & 0.6246 & 22.92\% & 0.7892 \\
\midrule
Neural Cali. ($\bD_{\tr}\rightarrow\bD_{\val}$) & 12.0\% & \bf 0.992 & \bf 4.59\% & \bf 0.7996 & \bf 10.91\% & \bf 0.7520 & \bf 3.66\% & \bf 0.6269 & \bf 22.87\% & \bf 0.7907 \\
\bottomrule
\end{tabular}
\label{table:ablation}\end{table*}

\subsection{Observing miscalibration}\label{sec:obs}
Here we would like to show some observations to demonstrate the issue of miscalibration, especially field-level miscalibration. These observations show that existing methods indeed have certain drawbacks in terms of specific metrics, which support the motivation and insight of our work.

{\it \textbf{Observation 1}: Adding more training data can help reach a higher AUC score, but does not indicate a smaller calibration error}.

It is observed in Table~\ref{table:loan}, the loan defaulter prediction task. We see that Model-2 significantly outperforms Model-1 in AUC, which is easy to understand as Model-2 is trained over more data. A machine-learning-powered decision-making system trained with more data tends to be more accurate. However, surprisingly, Model-2 suffered from higher calibration errors than Model-1 in all the calibration related metrics. We do not know where the error comes from, but we know that a model with such error is not reliable to use.

{\it \textbf{Observation 2}: Lower instance-level calibration error does not indicate lower field-level calibration error.}

This observation indicates the importance of our proposed field-level calibration errors. Table~\ref{table:criteo} shows an example. In this dataset, Model-2 not only gets higher AUC but also lower Log-loss and Brier score than the traditional post-hoc methods Isotonic Regression and Platt scaling. If only looking at these three metrics, it seems that Model-2 for this dataset is sufficient to use. However, when looking at our proposed field-level calibration errors, we find that Model-2 is worse in Field-ECE and Field-RCE, which means it is more biased in certain subsets of data than the calibrated models.

{\it \textbf{Observation 3}: Conventional calibration methods are sometimes better than Model-2 in calibration metrics, but worse in AUC.}

It can be observed across all the tested tasks. Particularly, Isotonic Regression and Platt Scaling did help improve the calibration metrics on dataset 1, 2, and 4, but failed on the third one and the last one. On the other hand, these calibration methods cannot help improving the AUC score over the base model, which is evident because they rely on an order-keeping function, so the AUC scores are always unchanged. Therefore, in AUC, traditional calibration methods are always worse than Model-2.

{\it \textbf{Observation 4}: Traditional post-hoc calibration methods are not reliable under data shift.}

For example, on the Avazu dataset in Table \ref{table:avazu}, the calibrated outputs of traditional methods get worse Field-RCE than the vanilla output. On the Lending Club data, Histogram Binning gets almost perfect Field-RCE but much worse AUC, indicating that it sacrifices the isotonic property to reduce the calibration error. These results show that when the data shift is undesirable, the univariate calibration methods are not so reliable.
It is also explained in a recent study about uncertainty estimation under data shift \cite{ovadia2019can} that these methods rely on the assumption that the data distribution of the development set is close to the test set. Otherwise, the calibrated output can be even worse than the uncalibrated outputs.

{\it \textbf{Observation 5}: Improvement in Field-level calibration errors is easier to observe and more interpretable than in instance-level metrics.}

We can see that the calibration methods often significantly reduce the field-level errors of Model-1. For example, in Table \ref{table:criteo}, calibration methods provided relative reductions on Field-ECE of about $20\%$ to $40\%$, and when using Neural Calibration, there is a significant Field-RCE reducuction from $7.46\%$ to $4.59\%$. However, the relative reduction in Log-loss and Brier score given by calibration methods are no more than 0.2\%, which is small enough to be ignored. Therefore, without looking at the field-level errors, one can neglect the miscalibration problem or underestimate the performance boosting given by calibration techniques.

\subsection{Main experimental results}\label{sec:expresults}
The main results are shown in Table \ref{table:loan}-\ref{table:tencent}. In each table, we show the base Model-1 as a baseline, Model-2 as a strong competitor against the calibration methods, two traditional univariate calibration methods, and our proposed Neural Calibration. We leave the results of Histogram Binning to Table \ref{table:ablation} due to the space limitation.

From the four columns in the middle of the tables, we found Neural Calibration the best in all calibration metrics, which is significantly better than all other tested methods in all tested datasets. Specifically, in our proposed metric Field-RCE, Neural Calibration is an order of magnitude smaller than the baseline Model-1 and Model-2 on the Lending Club dataset and the Porto Seguro dataset. These results are evidence that Neural Calibration can indeed reduce the error and make the probabilistic predictions more reliable.

Further, we see that Neural Calibration can get significantly higher AUC than conventional calibration methods, and reach comparable AUC with Model-2. Standard post-hoc calibration methods can not achieve such an improvement in AUC because they learn the univariate order-keeping mappings solely. In this way, we remark that Neural Calibration can make better use of the development dataset than previous methods.

\subsection{Ablation study: effectiveness of ILPS}
We tested the Isotonic Line-Plot Scaling solely to see if it is stronger than existing post-hoc methods that learn univariate mapping functions.
We run the calibration methods upon the logits provided by the base Model-1. We tested two conventional non-parametric methods, Histogram Binning and Isotonic Regression, and two parametric methods, Platt Scaling and our proposed ILPS.
The results are shown in the upper half of Table \ref{table:ablation}.

We see that ILPS consistently outperformed Platt scaling on all the datasets, and is comparable or better than non-parametric methods on dataset 2-5. In detail, we observe that the non-parametric methods outperform Platt Scaling in most tasks. The most straightforward method, Histogram Binning, is surprisingly good in calibration, but it sometimes hurts the AUC score because the mapping is not guaranteed to be non-decreasing. Isotonic Regression is robust in both Field-ECE and AUC across tasks, indicating that the non-decreasing property is necessary. So it is more reliable than Histogram Binning, although sometimes it is worse in the calibration metric. Overall, ILPS achieves considerably good results in both Field-ECE and AUC. Considering that it is also as efficient as any parametric methods if deployed in online learning, we think ILPS can be a good alternative to Platt Scaling.

\subsection{Ablation study: contribution of each part}
Recall that our Neural Calibration model consists of two modules: a univariate ILPS function $\eta(l)$ and a neural network $g(\bx)$. Now we would like to check the contribution of each part in terms of the two showed metrics, Field-RCE and AUC. The results related to this part are shown in the following rows in Table \ref{table:ablation}: Base ($\bD_{\tr}$), ILPS (parametric), Base ($\bD_{\val}$ only), and the last Neural Calibration row.

First, the ILPS part can already reach a good Field-RCE score stand-alone on all the data except for the Avazu dataset, even though it is designed to perform calibration on the probability-level. It seems that we do not necessarily need to use the field-aware calibration. But when we add the field-aware part, in the last row of the table, we see Neural Calibration achieved better field-level calibration than ILPS over all the tests. So it indicates that introducing the field-aware part does help calibration.

Second, ILPS itself cannot improve the AUC due to its order-keeping property. On the other hand, when a base model is trained on the development set only, it cannot reach a high AUC score either because the amount of data in the development set is too small. So both parts contribute to the high AUC scores of Neural Calibration: the ILPS part provides a baseline AUC score, and the auxiliary neural network improves upon it.
\subsection{Ablation study: contribution of the features}

Intuitively, when considering the field-level bias on a specific field $z$, one might think of training a model on the field $z$. For example, if an advertising system finds the actual click rate for an ad is 10\% lower than the system's estimation, it can calibrate it by multiplying the estimation by 90\% for the next time. So now we want to answer the following question: In Neural Calibration, why do we use all the input features $\bx$, instead of using only the sensitive field $z$?

We conduct an experiment by training a bivariate Neural Calibration model by replacing $\bx$ with $z$~(so it only involves two variables, $l$ and $z$). The results are shown in the ``Neural Cali (ILPS \& field $z$ only)'' row of Table \ref{table:ablation}.

It shows that this bivariate model is good at reducing the field-level calibration error on $z$. However, the improvements in AUC against the base model are limited, because the field $z$ only provides limited information. On the other hand, our Neural Calibration model trained with all the features can also achieve low Field-ECE comparable to the bivariate model and wins on the last two datasets. It means that introducing the features other than $z$ might not hurt the calibration performance on field $z$. Moreover, it can be inferred that Neural Calibration can also reduce the biases over other fields, which cannot be done by the bivariate model. Also, the full model is significantly better in AUC than the bivariate model. Therefore, we conclude that a Neural Calibration model trained with all the input features is empirically better than only using one field $z$.

\subsection{Ablation study: robustness under data shift}
From the observations mentioned previously, we notice that a key drawback of traditional post-hoc calibration methods is that they are not robust in the face of data shift. On the contrary, Neural Calibration works fairly good by adding the auxiliary neural net over all the datasets. For example, on the Avazu dataset, it reaches lower Field-RCE than all competitors.

Now we would like to see if the good performance of Neural Calibration is because the development set is closer to the test set. To show it, we set-up an ablation study by altering the development set and the training set, i.e., we first train the base model on the development set and then calibrate on the training set. The results are shown in the last 2 to 4 rows in Table \ref{table:ablation}.

In such an inverted setting, Neural Calibration is still surprisingly good. Again, we observe that in the Lending Club dataset, ILPS sacrifices AUC to fix miscalibration~(note that the isotonic constraint in ILPS is softened). But when the field-aware information is taken into account, both AUC and Field-RCE improved. It means that maybe some field accounts for the conflict between calibration and accuracy, so the field-aware module can fix it.

Moreover, from the experiment of inverted datasets, we see that Neural Calibration achieved high sample efficiency because every part of data is used to improve AUC or calibration. Even though it was first trained over a small dataset and calibrated over a dataset whose distribution is not so close to the test set, which seems improper in some sense, the performance was still considerably good. Thus we think our method is empirically reliable and robust.

\section{Conclusion and discussion}
This paper studied the issue of miscalibration for probabilistic predictions in binary classification. We first put forward the Field-level Calibration Error as a new class of metrics to measure miscalibration. It can report the biases on specific subsets of data, which is often overlooked by common metrics. Then we observed that existing post-hoc calibration methods have limited sample efficiency on the labeled data of the development set because they basically do probability-level calibration and cannot improve other non-calibration metrics such as the AUC score. So we proposed a simple yet powerful method based on neural networks, named Neural Calibration, to address this issue. It consists of 1) a novel parametric calibration mapping named Isotonic Line-Plot Scaling, which is non-decreasing, continuous, and has strong fitting power, and 2) an auxiliary neural network for field-aware calibration. We tested our method on five large-scale datasets. By using the pipeline of Neural Calibration, we achieved significant improvements over conventional methods. Specifically, the parametric ILPS alone outperformed Platt Scaling on all the calibration metrics, and Neural Calibration can further improve the AUC score. Also, we found our method robust in the face of dataset shift. Thus we conclude that the proposed field-aware calibration can achieve high performance on both calibration and non-calibration metrics simultaneously and is reliable for practical use in real-world applications.

We think it is a promising direction for the future work to understand the source of miscalibration error in modern machine learning, especially in deep learning and information retrieval systems. By considering other types of problem settings, future work includes extending field-aware calibration into regression and multi-class classification, or extending the supervised learning setting to more general settings online learning and reinforcement learning.

\section*{acknowledgements}
Special thanks to Zhixiang (Plantsgo) Hua for insightful discussions.
This work is supported by the National Key Research and Development Program of China under Grant No. 2017YFB1002104, National Natural Science Foundation of China under Grant No. 61976204, U1811461, 91846113, Project of Youth Innovation Promotion Association CAS. It is also supported by CCF-Tencent RhinoBird Young Faculty Open Research Fund No. RAGR20180111 and 2019 Tencent Marketing Solution Rhino-Bird Focused Research Program.

\bibliographystyle{ACM-Reference-Format}
\bibliography{cali}

%
%
%
%
%
%
%
%
%
\end{document}